\documentclass[11pt]{article}

\usepackage[final]{acl}

\usepackage{times}
\usepackage{latexsym}

\usepackage[T1]{fontenc}

\usepackage[utf8]{inputenc}

\usepackage{microtype}

\usepackage{times}
\usepackage{latexsym}
\usepackage{subcaption}
\usepackage{multirow}
 \usepackage{amsmath}
\usepackage{booktabs}

 \usepackage[flushleft]{threeparttable}
 
\usepackage{inconsolata}

\usepackage{graphicx}

\usepackage{shellesc}

\usepackage{textcomp}  
\usepackage{scalerel}  
\usepackage[outdir=latex]{epstopdf}

\def\pleading{\hspace{0.5mm}\raisebox{-0.7ex}{\scalerel*{\includegraphics{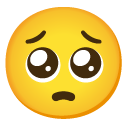}}{\mbox{\huge X}}}\hspace{1mm}}

\def\vomiting{\hspace{0.5mm}\raisebox{-0.7ex}{\scalerel*{\includegraphics{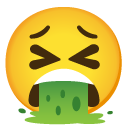}}{\mbox{\huge X}}}\hspace{1mm}}

\def\tear{\hspace{0.5mm}\raisebox{-0.7ex}{\scalerel*{\includegraphics{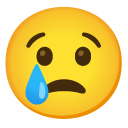}}{\mbox{\huge X}}}\hspace{1mm}}

\def\smiletear{\hspace{0.5mm}\raisebox{-0.7ex}{\scalerel*{\includegraphics{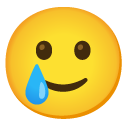}}{\mbox{\huge X}}}\hspace{1mm}}

\def\smilingfacewithhearts{\hspace{0.5mm}\raisebox{-0.7ex}{\scalerel*{\includegraphics{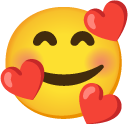}}{\mbox{\huge X}}}\hspace{1mm}}

\def\nomouth{\hspace{0.5mm}\raisebox{-0.7ex}{\scalerel*{\includegraphics{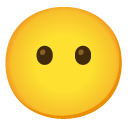}}{\mbox{\huge X}}}\hspace{1mm}}

\def\scream{\hspace{0.5mm}\raisebox{-0.7ex}{\scalerel*{\includegraphics{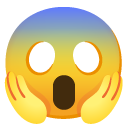}}{\mbox{\huge X}}}\hspace{1mm}}

\def\hug{\hspace{0.5mm}\raisebox{-0.7ex}{\scalerel*{\includegraphics{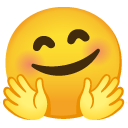}}{\mbox{\huge X}}}\hspace{1mm}}

\def\thinking{\hspace{0.5mm}\raisebox{-0.7ex}{\scalerel*{\includegraphics{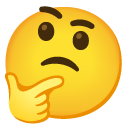}}{\mbox{\huge X}}}\hspace{1mm}}

\def\confused{\hspace{0.5mm}\raisebox{-0.7ex}{\scalerel*{\includegraphics{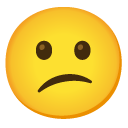}}{\mbox{\huge X}}}\hspace{1mm}}

\def\raisehand{\hspace{0.5mm}\raisebox{-0.7ex}{\scalerel*{\includegraphics{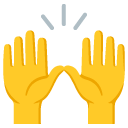}}{\mbox{\huge X}}}\hspace{1mm}}

\def\joy{\hspace{0.5mm}\raisebox{-0.7ex}{\scalerel*{\includegraphics{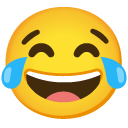}}{\mbox{\huge X}}}\hspace{1mm}}

\def\grinningsmile{\hspace{0.5mm}\raisebox{-0.7ex}{\scalerel*{\includegraphics{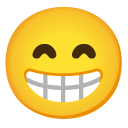}}{\mbox{\huge X}}}\hspace{1mm}}

\def\hearteyes{\hspace{0.5mm}\raisebox{-0.7ex}{\scalerel*{\includegraphics{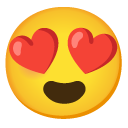}}{\mbox{\huge X}}}\hspace{1mm}}

\def\grimacing{\hspace{0.5mm}\raisebox{-0.7ex}{\scalerel*{\includegraphics{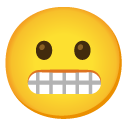}}{\mbox{\huge X}}}\hspace{1mm}}

\def\sad{\hspace{0.5mm}\raisebox{-0.7ex}{\scalerel*{\includegraphics{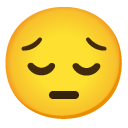}}{\mbox{\huge X}}}\hspace{1mm}}

\def\dancer{\hspace{0.5mm}\raisebox{-0.7ex}{\scalerel*{\includegraphics{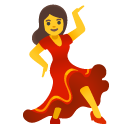}}{\mbox{\huge X}}}\hspace{1mm}}

\def\thumbsup{\hspace{0.5mm}\raisebox{-0.7ex}{\scalerel*{\includegraphics{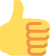}}{\mbox{\huge X}}}\hspace{1mm}}

\def\smirk{\hspace{0.5mm}\raisebox{-0.7ex}{\scalerel*{\includegraphics{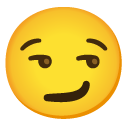}}{\mbox{\huge X}}}\hspace{1mm}}

\def\fist{\hspace{0.5mm}\raisebox{-0.7ex}{\scalerel*{\includegraphics{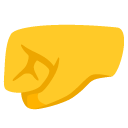}}{\mbox{\huge X}}}\hspace{1mm}}

\def\question{\hspace{0.5mm}\raisebox{-0.7ex}{\scalerel*{\includegraphics{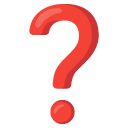}}{\mbox{\huge X}}}\hspace{1mm}}

\def\wink{\hspace{0.5mm}\raisebox{-0.7ex}{\scalerel*{\includegraphics{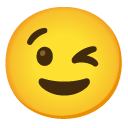}}{\mbox{\huge X}}}\hspace{1mm}}

\usepackage{xcolor}
\usepackage{soul}

\title{The Prosody of Emojis}

\author{
  Giulio Zhou$^{1}$\thanks{Corresponding author: \texttt{giulio.zhou@ed.ac.uk}} \quad\quad
  Tsz Kin Lam$^{2}$\thanks{Work done while at the University of Edinburgh.} \quad\quad
  Alexandra Birch$^{1}$ \quad\quad
  Barry Haddow$^{3}$\footnotemark[2] \\
  \\
  $^{1}$University of Edinburgh \quad $^{2}$NatWest \quad $^{3}$Aveni
}

\begin{document}
\maketitle
\begin{abstract}
Prosodic features such as pitch, timing, and intonation are central to spoken communication, conveying emotion, intent, and discourse structure. In text-based settings, where these cues are absent, emojis act as visual surrogates that add affective and pragmatic nuance. This study examines how emojis influence prosodic realisation in speech and how listeners interpret prosodic cues to recover emoji meanings. Unlike previous work, we directly link prosody and emojis by analysing human speech data collected through a controlled elicited production task\footnote{\url{https://huggingface.co/datasets/GiulioZh/ProsodyEmoji}}. Using Bayesian multilevel modelling, we show that speakers systematically adapt their prosody based on emoji cues, and that listeners can recover intended meanings significantly above chance. Furthermore, our results reveal a clear hierarchy in prosodic shifts: greater semantic differences between emojis correspond to increased prosodic divergence. These findings suggest that emojis are meaningful carriers of prosodic intent that bridge the gap between digital text and spoken production.

\end{abstract}

\section{Introduction}

\begin{figure}[t]
\centering
  \includegraphics[width=7.8cm]{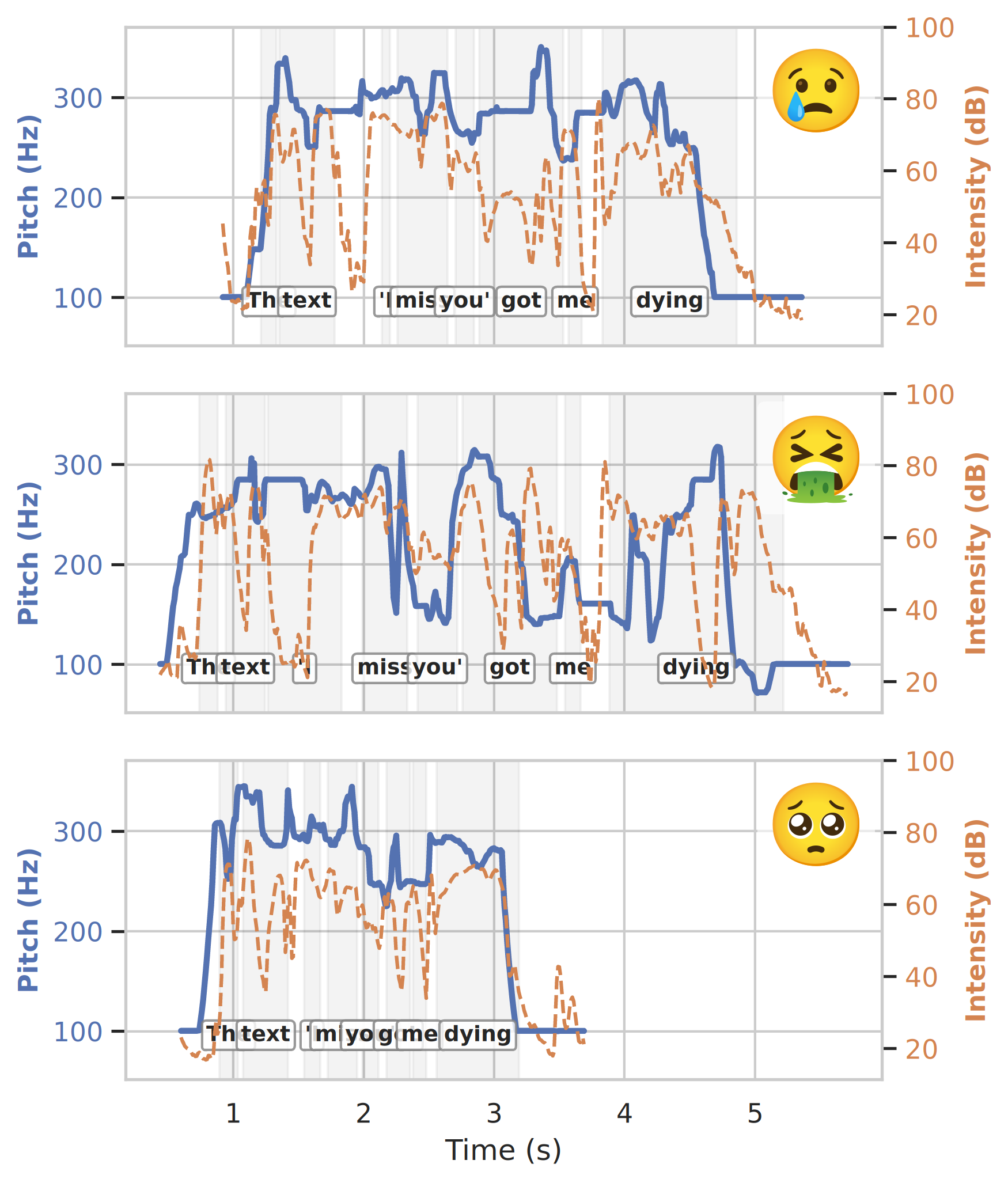}
  \caption{Prosodic variations for the sentence \textit{“The text `I miss you' got me dying”} across three emoji contexts. Each subplot displays the pitch (solid blue, Hz) and intensity (dashed orange, dB) contours. }
  \label{fig:example}
\end{figure}

In spoken language, prosody, encompassing pitch, rhythm, intonation, and other paralinguistic features, plays a crucial role in conveying linguistic nuances and socio-emotional cues \citep{cole2015prosody, hellbernd2016prosody, ward-levow-2021-prosody}. Acting as a dynamic carrier of information, prosody enriches spoken communication by modulating affect, signalling discourse structure, and shaping listener expectations. However, the shift toward computer-mediated communication presents challenges: text-based interaction strips away vital non-verbal elements like intonation, gesture, and facial expression, essential to holistic interpretation \citep{archer1977words, Crystal_2001}.

To compensate for this loss, users of digital communication platforms increasingly rely on emojis, which enrich digital communication by offering a visual medium for expressing emotions \citep{gulcsen2016you} and conveying semantic meanings \citep{naaman-etal-2017-varying}, transcending textual constraints. However, their interpretive flexibility suggests a functional overlap with the paralinguistic channel. Given that prior research confirms prosodic features cannot be fully recovered from text alone \citep{wolf-etal-2023-quantifying}, a critical question emerges: can emojis, embedded in otherwise static written language, serve as mediators for prosodic intent and compensate for the absence of vocal cues? Specifically, we investigate the following research questions:

\vspace{3mm} 
{\textbf{RQ1:} Do emojis influence how prosodic features are expressed in spoken language?}    

{\textbf{RQ2:} Can listeners interpret prosodic cues to recover intended emoji meanings in speech?}    

{\textbf{RQ3:} How does inter-speaker similarity in prosodic realisation relate to listener interpretation of emoji meaning?}

{\textbf{RQ4:} Is there a correlation between the semantic distance of emojis and the magnitude of the prosodic shift in their vocal delivery?}

\vspace{3mm}

To investigate these questions, we conducted four online experiments to collect a dataset of emoji-enriched speech and listener perception data (Figure \ref{fig:example}). We employed a design that holds lexical content constant across emoji conditions to isolate the influence of emojis on prosody, thereby preventing confounding effects from varying syntax or vocabulary.
While this involves reading specific sentences rather than recording spontaneous conversations, we do not instruct participants on how to deliver the text or focus on the emojis. This approach captures a broad range of naturalistic expressive strategies while ensuring that observed prosodic shifts reflect authentic interpretations of emoji meaning.

Our results demonstrate that emojis consistently invite prosodic modulation across diverse speakers, even with constant verbal content. This variation is often sufficient for listeners to recover intended emoji meanings significantly above chance. We find that prosodic convergence, where different speakers independently produce similar contours for the same emoji, significantly boosts listener success. This suggests that shared prosodic intuitions facilitate robust interpretation. Finally, our analysis reveals a hierarchy in prosodic shifts. The magnitude of prosodic change correlates with emoji semantic distance, as greater semantic dissimilarity between emojis leads to systematically larger prosodic divergence.

\section{Background}

\subsection{Emoji Interpretation}

Emojis constitute a pervasive form of digital paralanguage: graphical cues that represent the non-verbal behaviours, such as facial expressions and gestures, intrinsic to face-to-face interaction \citep{alshenqeeti2016emojis}. From a psychological perspective, emojis do not merely decorate text but actively modulate affective processing. They have been shown to trigger cognitive responses in the receiver similar to those elicited by physical facial expressions, thereby influencing social attributions of the sender, such as perceived warmth, trust, and sincerity, and promoting a sense of social presence \citep{boutet2021emojis}.

Beyond their social impact, emojis serve critical pragmatic functions by acting as illocutionary force indicators. They allow users to clarify intent and manage face-threatening acts, such as softening a critique or signalling irony, which are tasks typically reserved for vocal nuance \citep{li2018pragmatic, holtgraves2020emoji}. However, this interpretive process is fraught with ambiguity. While literal meanings are relatively stable across languages, figurative meaning is closely intertwined with the sentiment of the context \citep{zhou-etal-2024-semantics}. Despite the presence of contextual cues intended to clarify intent, empirical evidence shows that misinterpretation persists, suggesting that emojis remain a high-variance channel that context alone cannot resolve \citep{miller2017understanding, czkestochowska2022context}.

\subsection{Prosody in Speech Processing}

Prosody encompasses the suprasegmental features of speech that operate beyond individual phonemes to structure discourse and convey speaker intent \cite{lehiste1976suprasegmental, gerken1998overview}. Despite the theoretical importance of prosody, speech processing systems have historically prioritised lexical content, particularly in cascade pipelines where prosodic nuance is stripped away during transcription \cite{zhou-etal-2024-prosody, tsiamas-etal-2024-speech}. However, recent research underscores that prosodic cues are essential for robust computational understanding, independently supporting complex downstream tasks such as spoken question-answering, emotion recognition, and speaker profiling \cite{hirschberg2002communication, chi-etal-2025-role}. Beyond understanding, prosody is also critical for speech generation. In Text-to-Speech (TTS) synthesis, naturalness and listener engagement heavily depend on accurate prosodic modelling and expressive transfer \cite{skerry2018towards, kane2024voice}.

\subsection{Prosody and Emojis}

While computational models struggle to generate natural prosody from plain text, human readers actively construct speech rhythm and tone using visual cues. Punctuation guides implicit prosodic boundaries during silent reading \cite{drury2016punctuation}, while typographic markers like font weight and baseline shifts provide explicit instructions for vocal intensity, pitch, and affective expression \cite{de2020speech}.

Within this framework, emojis function as an evolutionary extension of these traditional typographical cues. Given that emojis serve as pragmatic substitutes for non-verbal behaviour, they present a unique opportunity to recover the prosodic intent lost in digital text. Prior research has drawn conceptual parallels here, likening emojis to written intonation or visual facial expressions \cite{james2017prosody} that highlight information focus or compensate for a lack of vocal expressivity \cite{wagner2016kind, hu2019emojilization, kaiser2021focus}.
Despite these theoretical links, empirical analysis of how emojis influence the acoustic realisation of speech remains limited. While some work has proposed emojis as graphical surrogates for prosody based on surveys \cite{alnuzaili2024emojis}, or used emoji-enriched speech to increase the variance of TTS systems \cite{tuttosi2025emojivoice}, research has yet to characterise the systematic prosodic adaptations human speakers make in response to different emoji semantics. In contrast, our study examines how speakers naturally modulate their acoustic prosody when encountering emoji-enriched utterances. By analysing real human speech, we provide empirical evidence that emoji semantics directly influence prosodic expression, bridging the gap between symbolic digital cues and spoken production.

\section{Methods}

\subsection{Data Collection}
To investigate our research questions, we conducted four online experiments\footnote{See Appendix \ref{app:instructions} for full instructions and Appendix \ref{app:data} for data summaries.}, with each experiment employing a unique pool of participants. To isolate the influence of emojis on prosody, we employed a design that holds lexical content constant across conditions. While this involves elicited speech rather than spontaneous conversation, this approach is necessary to generate the parallel data required to compare how the same lexical string is transformed by different emoji contexts, a phenomenon rarely captured in naturalistic corpora.

The experimental framework is built upon a base corpus of 230 unique utterances, each of which originally contained at least one emoji. These were manually selected and edited from publicly available user-generated textual content to cover a broad range of topics (e.g., music, gaming, film) and diverse emoji usage. All selected utterances were specifically screened to ensure they remained natural when spoken aloud.

\subsubsection*{Experiment 1: Utterance-Emoji Stimulus}

The objective of Experiment 1 was to construct a collection of utterances with identical lexical content but varying emoji usage. Participants were presented with the base utterances (with original emojis removed) and instructed to create up to five variations per sentence by inserting different emojis to alter the emotional tone or interpretation. To ensure variety, they were advised against using emojis with highly similar meanings for the same utterance. Each utterance was annotated independently by two annotators, resulting in 1,595 unique emoji-augmented stimuli.

\subsubsection*{Experiment 2: Emoji-Influenced Prosodic Data}

We collected a corpus of emoji-influenced speech by asking participants to read a subset of the stimuli annotated in Experiment 1. Each speaker recorded approximately 25 utterances, ensuring they produced at least two distinct variations of each base sentence (including a non-emoji control version). To minimise bias in prosodic expression, participants were not informed of the focus on emojis. Following the recordings, participants summarised the perceived meaning of each emoji in one word to verify their engagement and interpretation. The final dataset comprised 3,153 valid recordings from 129 British English speakers.
To account for potential sources of variance in our analysis, we also collected demographic data, including age, gender, and self-reported general emoji usage (measured on a scale of 1–5). Additionally, participants reported their recording hardware (headset, mobile device, or built-in computer microphone).

\subsubsection*{Experiment 3: Judging Prosodic Variation Within Speakers}

To test listeners' perception of prosodic changes within the same speaker, we sampled 645 recording pairs which had identical lexical content. These pairs represented two distinct contrast types: \textit{Emoji--Emoji}, comparing two distinct emoji renditions (e.g., \emph{Text} + \sad vs. \emph{Text} + \smilingfacewithhearts), and \textit{Emoji--NoEmoji}, comparing an emoji rendition against an emoji-free control. A total of 107 unique listeners participated in this task. For each pair, listeners judged whether the recordings differed in prosodic realisation (RQ1) and matched each recording to its corresponding emoji (RQ2). Listeners were instructed to focus on vocal expressiveness and ignore recording artefacts. Each pair received judgments from an average of three annotators. Consistent with Experiment 2, we collected demographic data from these listeners, including age, gender, and general emoji usage (scale 1-5).

\subsubsection*{Experiment 4: Identifying Shared Prosody Across Speakers}

Experiment 4 investigated whether inter-speaker similarity in prosodic realisation, or prosodic convergence, facilitates the interpretation of emoji meaning (RQ3). To ensure sufficient expressive range, we excluded speakers who demonstrated low prosodic variation in Experiment 3 (see Section \ref{sec:speakervariation}). The final stimulus set comprised 49 inter-speaker audio pairs, each consisting of recordings of the same sentence and emoji produced by two different speakers. A total of 10 unique listeners evaluated these pairs. Participants performed two tasks: judging the prosodic similarity of the recordings, and matching each recording to either the target emoji or a plausible distractor drawn from a curated list of 60 varied face emojis. Crucially, while both recordings in a pair shared the same target emoji, participants were not informed of this fact. Each pair received judgments from an average of four annotators. As in previous experiments, we collected demographic data from these listeners, including age, gender, and general emoji usage.

\subsection{Statistical Analysis}

To evaluate our research questions, we implemented Bayesian linear mixed-effects regression models (Appendix \ref{app:models}). This framework allows us to directly quantify the strength of evidence for our hypotheses by estimating the full probability distribution of our parameters. We determined our model structures and likelihood functions \textit{a priori} based on our experimental design, modelling binary perception outcomes (RQ1--RQ3) using Bernoulli distributions and continuous acoustic distances (RQ4) using Gaussian distributions. By incorporating random effects for speakers, listeners, and specific utterances, we ensured that our population-level estimates account for individual expressive variability and inherent prosodic differences between sentences. Detailed model specifications including prior choices, convergence checks, and adherence to the maximal random effects principle \citep{barr2013random} are provided in Appendix \ref{app:models}.

\section{Prosody in Emoji-Enriched Speech}

\subsection{Speaker Variation in Prosodic Expression}\label{sec:speakervariation}

\begin{figure}[t]
\centering
  \includegraphics[width=\linewidth]{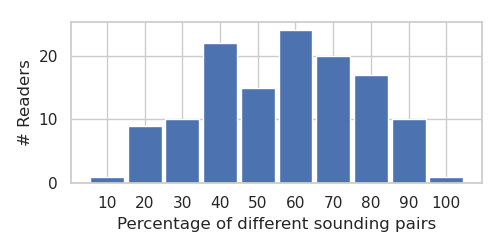}
  \caption{Readers (Experiment 2) grouped by the proportion of recordings judged as prosodically different (Experiment 3). Each bin represents a 10\% range (e.g., 0\%–9\%), from least to most prosodically varied speakers. }
  \label{fig:stats}
\end{figure}

\begin{table}[t]
\centering
\resizebox{\columnwidth}{!}{
\begin{tabular}{lrc}
\toprule
\textbf{Predictor} & \textbf{Est. ($\beta$)} & \textbf{95\% CI} \\
\midrule
(Intercept) & -0.25 & [-0.66, 0.15] \\ 
Age & -0.04 & [-0.16, 0.07] \\ 
Gender (Female) & 0.16 & [-0.12, 0.44] \\
Mic (Headset) & 0.10 & [-0.26, 0.46] \\
Mic (Mobile) & -0.15 & [-0.42, 0.12] \\
Emoji Use & \textbf{0.18} & \textbf{[0.06, 0.30]} \\ 
Contrast Type (Emoji--Emoji) & \textbf{0.29} & \textbf{[0.03, 0.55]} \\ \bottomrule
\end{tabular}}
\caption{Model 1 results for perceived prosodic differentiation. Bold indicates 95\% Credible Intervals that strictly exclude zero.}
\label{tab:speaker_variation}
\end{table}

Our first inquiry examines whether emojis elicit prosodic variation (RQ1). We analysed data from Experiment 3, where listeners compared pairs of recordings produced by the same speaker. Figure \ref{fig:stats} shows the number of speakers grouped by the proportion of their recording pairs judged as prosodically distinct. While 20 speakers exhibited minimal differentiation (0–30\%), the majority (81 speakers) showed moderate variation (31–70\%), and a notable subset (28 speakers) consistently produced distinct renditions (71–100\%). To quantify this observed variance and test the overall effect of emojis, we modelled listener judgments using the following Bayesian multilevel logistic regression:

\begin{center}\textbf{Model 1}: \texttt{PerceivedVariation $\sim$ Age + Gender + Mic + EmojiUse + ContrastType + \\(1 | Utterance) + (1 | Listener)}\end{center}

In this model, \texttt{PerceivedVariation} is the binary response variable (whether the two recordings sounded different or not), while self-reported \texttt{EmojiUse}, speaker demographics (age, speaker gender, microphone type), and \texttt{ContrastType}, which accounts for whether the listener was comparing a pair of recordings with two different emojis (Emoji--Emoji) or an emoji against an emoji-free control (Emoji--NoEmoji), serve as the fixed effects.

The results, detailed in Table \ref{tab:speaker_variation}, provide a nuanced answer to RQ1. The influence of emojis appears to be a capacity that varies significantly between individuals. While most speakers exhibit moderate variation, there is significant spread, with clear tails of highly expressive and highly monotonous speakers (see Section \ref{sec:speakervariation}). Model 1 helps explain this variance through \texttt{EmojiUse} ($\beta = 0.18, CI [0.06, 0.30]$). The positive effect suggests that for frequent emoji users, the parallel between emoji semantics and prosody is stronger, leading to more distinct acoustic signals. Furthermore, \texttt{ContrastType} systematically influenced detectability. The positive coefficient for Emoji--Emoji trials ($\beta = 0.29, CI [0.03, 0.55]$) indicates that listeners were significantly more likely to detect a difference when comparing two distinct emojis than when evaluating an Emoji--NoEmoji pair. This implies that emojis do not simply trigger a generic layer of extra expressiveness. Instead, speakers adapt their prosody to the specific meaning of each emoji. Because these adaptations pull the vocal delivery in distinct, divergent directions, the acoustic gap between two different emojis is naturally wider, and easier for listeners to hear, than the gap created by simply adding one emoji to a text. These findings verify the existence of the parallel proposed in RQ1: emojis can and do influence prosodic expression. However, this influence is not uniform. Instead, emojis function as an expressive resource that is modulated by the speaker's digital fluency and the specific semantic context. 

\subsection{Listener Interpretation of Prosodic Variation}\label{sec:ex2res}

\begin{table}[t]
\centering
\resizebox{\columnwidth}{!}{
\begin{tabular}{lrc}
\toprule
\textbf{Predictor} & \textbf{Est. ($\beta$)} & \textbf{95\% CI} \\
\midrule
\multicolumn{3}{l}{\textit{Model 2: Intra-speaker Identification (RQ2)}} \\
(Intercept) & -0.16 & [-0.45, 0.11] \\
Speaker Expressivity & \textbf{0.24} & \textbf{[0.10, 0.38]} \\
Contrast Type (Emoji--Emoji) & 0.03 & [-0.27, 0.32] \\
Speaker Emoji Use & 0.05 & [-0.09, 0.19] \\
Listener Emoji Use & -0.15 & [-0.31, 0.01] \\
Speaker$\times$Listener Use & 0.09 & [-0.05, 0.23] \\
\midrule \multicolumn{3}{l}{\textit{Model 3: Inter-speaker Identification (RQ3)}} \\
(Intercept) & -1.79 & [-3.38, -0.17] \\ Perceived Similarity (Yes) & \textbf{2.42} & \textbf{[1.66, 3.22]} \\ \bottomrule
\end{tabular}}
\caption{Model 2 (intra-speaker) and Model 3 (inter-speaker) results for emoji identification success. Bold indicates 95\% Credible Intervals that strictly exclude zero.}
\label{tab:combined_identification}
\end{table}

Beyond mere production, we investigated whether these variations effectively communicate symbolic intent and allow listeners to recover intended emoji meanings (RQ2). We tested this using a second model:
$$$$
\begin{center}
     \textbf{Model 2}: \texttt{IdSuccess $\sim$ SpeakerExpressivity + ContrastType + \ (SpeakerEmojiUse $\times$ ListenerEmojiUse) + \ (1 | Utterance) + (1 | Listener)}
\end{center}

In this model, \texttt{IdSuccess} denotes the correct assignment of emoji meanings to both recordings in the pair. Key predictors included \texttt{SpeakerExpressivity} (a proxy for expressive bandwidth derived from Experiment 3), \texttt{ContrastType} (Emoji--Emoji vs Emoji--NoEmoji), and the interaction between speaker and listener \texttt{EmojiUse} to test for shared digital fluency effects. Model 2 (Table \ref{tab:combined_identification}) revealed a baseline identification probability of $P(\text{IdSuccess}) \approx 0.46$ (Intercept = $-0.16$). Although the intercept coefficient includes zero (indicating a probability near 50\%), this value sits significantly above the 25\% chance level\footnote{Chance performance is 25\% based on the four possible emoji assignment combinations available to the listener (Log-odds $\approx -1.1$).}, confirming that the task was generally performable.

The primary driver of communicative success was Speaker Expressivity: for every standard deviation increase in a speaker's prosodic differentiation, the log-odds of correct emoji identification increased reliably ($\beta = 0.24, CI [0.10, 0.38]$). Interestingly, while Contrast Type was a major factor in the perception of differences (Model 1), it did not reliably impact identification success in Model 2 ($\beta = 0.03, CI [-0.27, 0.32]$). This suggests that once a prosodic difference is detected, the intended meaning is equally recoverable whether comparing two different emojis or an emoji against an emoji-free control. Additionally, the lack of a reliable interaction between speaker and listener emoji familiarity indicates that this communication is mutually intelligible regardless of digital background, relying instead on universal expressive mechanisms. These findings address RQ2 by demonstrating that while listeners can interpret prosodic cues to recover emoji meanings significantly above chance, the success of this communication is not universal; rather, it is a high-variance channel that depends largely on the individual speaker's ability to encode pragmatic intent into vocal variation.

\subsection{Prosodic Convergence and Emoji Interpretation}

RQ3 shifts focus to the inter-speaker dynamics explored in Experiment 4. Specifically, we investigated whether prosodic convergence, defined as instances where different speakers independently produce similar contours for the same emoji, facilitates higher listener accuracy. To test this, we modelled the relationship between the listener's perception of similarity and their ability to correctly identify the emoji using a third Bayesian model:

\begin{center}

\textbf{Model 3}: \texttt{IdSuccess $\sim$ PerceivedSimilarity + \ (1 | Utterance) + (1 | Listener) + \\(1 | Speaker$_A$) + (1 | Speaker$_B$)}
\end{center}

The fixed effect \texttt{PerceivedSimilarity} is a binary predictor indicating whether the listener judged the recordings from the two speakers as sounding prosodically similar. Random effects account for variance across utterances, listeners, and the two unique speakers involved in each pair (\texttt{Speaker$_A$} and \texttt{Speaker$_B$}). The regression results, presented as Model 3 in Table \ref{tab:combined_identification}, confirm that perceived similarity is a robust predictor of successful emoji recovery. The model found a significant positive effect for similarity ($\beta = 2.42, 95\% \text{ CI } [1.66, 3.22]$). This result suggests that listeners are approximately 11 times more likely to achieve discriminative success when speakers converge on a shared prosodic signature. Even after controlling for random variation, the evidence for this effect remained extreme with a posterior probability of $1.0$. These findings provide a direct answer to RQ3 by demonstrating that inter-speaker similarity in prosodic realisation is fundamentally tied to listener interpretation success. When speakers independently converge on a shared prosodic strategy, listener recovery of intended meaning improves significantly. However, communicative success is notably more fragile when prosodic strategies diverge. In such cases, listeners often perceive the recordings as sounding different and prefer to assign distinct emoji meanings to each, reflecting the increased uncertainty caused by a lack of prosodic agreement (further details in Appendix \ref{app:data4}). Thus, while emoji vocalisation reflects a stable and shared paralinguistic code that can ensure semantic intent remains recoverable across different voices, its effectiveness is largely contingent on whether speakers align their prosodic intuitions.

\section{Prosodic Distance and Emoji Semantic Variations}

This section investigates how emoji semantics influence prosodic variation. To address RQ4, we test the hypothesis that prosodic differences are greater when associated emojis belong to distinct semantic categories than when they share a category. Additionally, we examine whether the inclusion of an emoji in a previously emoji-free sentence induces greater prosodic change than switching between emojis within the same semantic group.

\begin{figure}[h]
\centering
  \includegraphics[width=0.95\linewidth]{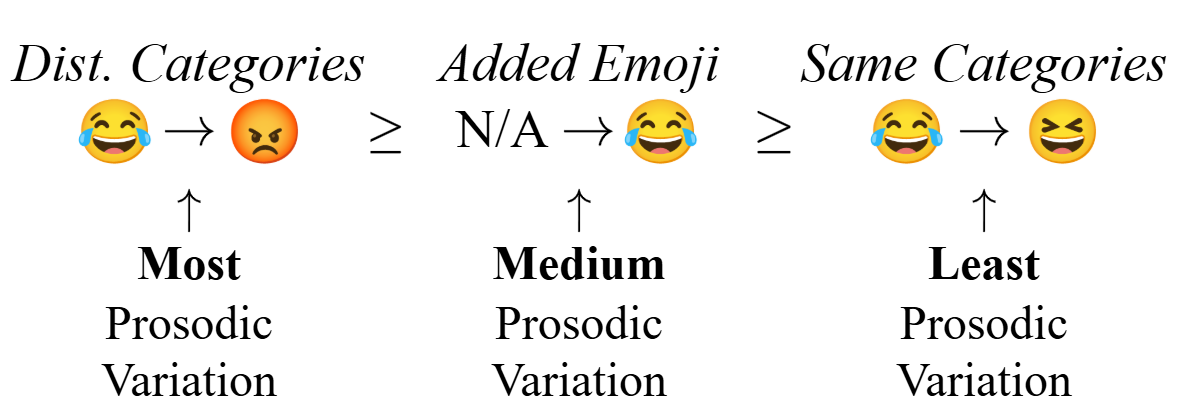}
  \label{fig:rq3hyp}
\end{figure}

\subsection{Emoji Semantic Categories}

\begin{figure}[t]
\centering
  \includegraphics[width=\linewidth]{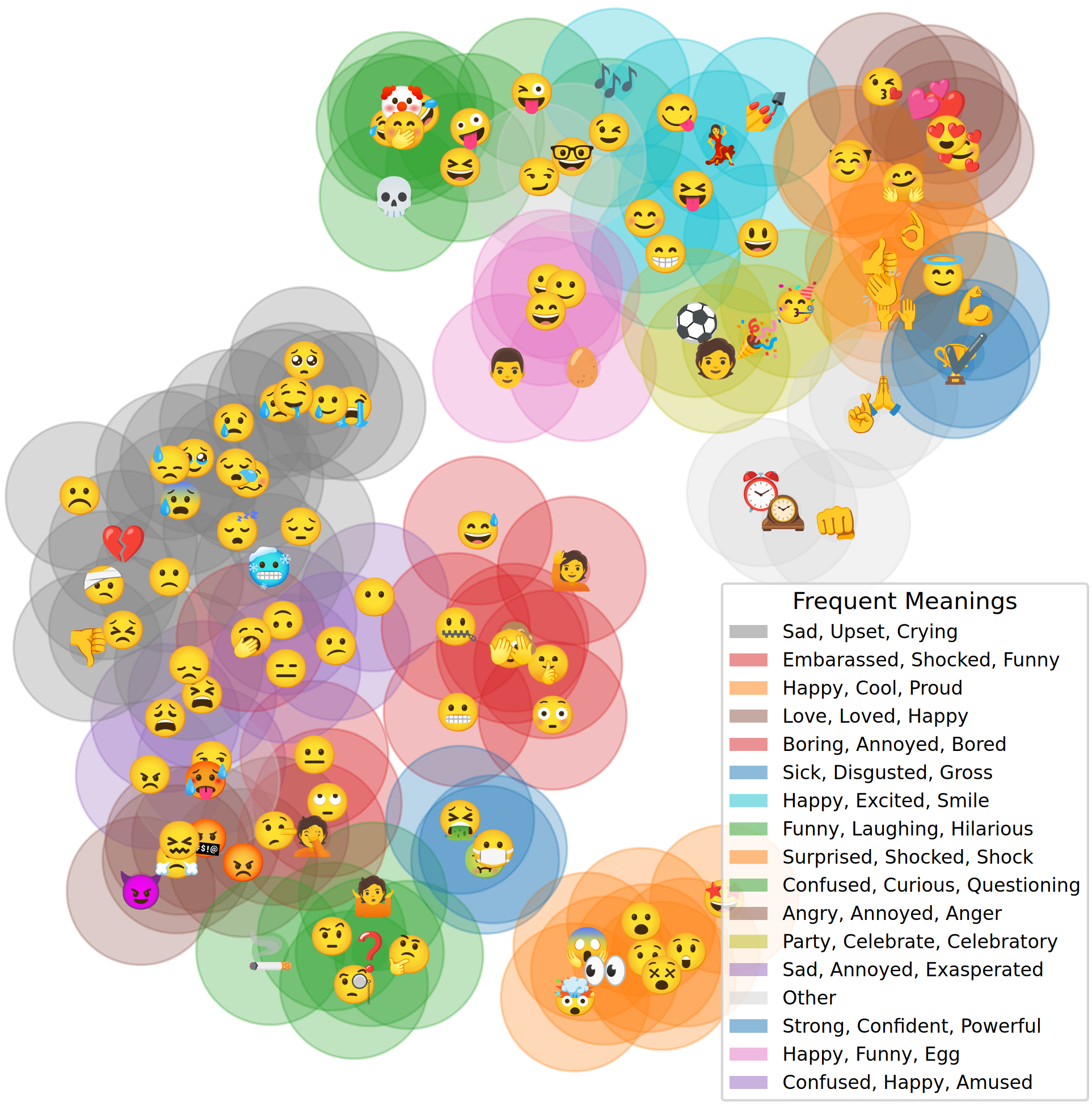}
  \caption{Emoji embeddings reduced with UMAP. Clusters computed with HDBSCAN (on 49 components). Labels are top 3 keywords among the words used to describe the emojis.}
  \label{fig:emoji_embeddings_ex2}
\end{figure}
Accurately categorising emojis poses a challenge because standard groupings, such as Unicode emoji groups and subgroups\footnote{\url{https://unicode.org/Public/emoji/16.0/emoji-test.txt}}, do not always reflect the communicative or semantic similarities perceived by users (e.g., \smiletear and \smilingfacewithhearts both belong to the \textit{“face-affection”} subgroup). To address this, we derived custom emoji embeddings from the free-text participant intents (Experiment 1) and interpretations (Experiment 2). We mapped these textual descriptions into a continuous semantic space using a pre-trained sentence transformer (\texttt{all-mpnet-base-v2} \cite{reimers-gurevych-2019-sentence}). Despite individual variation, these embeddings showed strong correlations between intents and interpretations, indicating a substantial shared understanding of the communicative roles of emojis (see Appendix \ref{app:embedding} for detailed analysis).

To identify meaningful semantic groups, for example, automatically grouping emojis described as “laughing” or “amused” into a single cluster, we performed unsupervised density-based clustering on the interpretation embeddings using HDBSCAN \cite{campello2013density} (Figure \ref{fig:emoji_embeddings_ex2}, see Appendix \ref{app:clustering} for detailed clustering parameters and evaluation). Based on these data-driven clusters, we established three experimental conditions for our prosodic analysis: recording pairs where the two emojis belong to different semantic clusters (Distinct Category); recording pairs where the emojis belong to the same semantic cluster (Same Category); and pairs consisting of an emoji rendition matched against an emoji-free recording (Added Emoji).

All relevant recording pairs were extracted and categorised accordingly, excluding speakers with minimal prosodic differentiation (see Section \ref{sec:speakervariation}).

\subsection{Methodology}
For each recording, we extracted multiple prosodic features to capture variation across several dimensions. Pitch contours were estimated using the FCNF0++ model \citep{morrison2023cross}, and intensity was calculated via the Praat Parselmouth interface \citep{parselmouth}. To account for varying recording lengths, we employed Dynamic Time Warping (DTW) to align pitch and intensity contours. From these paths, we derived a Temporal Warping metric defined as the percentage of non-diagonal steps in the alignment to quantify rhythmic divergence and local temporal distortion. Additionally, we calculated the absolute difference in speech rate (characters per second) as a measure of global articulation speed via the Stopes toolkit \citep{seamless}.

Beyond raw acoustic features, we extracted prosodic embeddings from the 7th layer of the Masked Prosody Model \citep{Wallbridge2025}. These frame-wise, 256-dimensional representations are explicitly trained to model pitch, energy, and voice activity. Distances between these sequences were computed using multivariate DTW to capture holistic prosodic divergence across these latent dimensions.

To determine if the magnitude of the prosodic shift varies systematically according to emoji semantic relationships (RQ4), we employed Bayesian linear mixed-effects regression. We fitted separate models for each acoustic feature using the following specification:

\begin{center}\textbf{Model 4}: \texttt{ProsodicDistance $\sim$ SemanticRelation + (1 + SemanticRelation | Speaker) + (1 | Utterance)}\end{center}

The dependent variable, \texttt{ProsodicDistance}, represents the calculated acoustic difference between the recordings. To satisfy normality assumptions and facilitate direct comparison across features, all distance metrics were log-transformed and Z-standardised prior to fitting. The fixed effect, \texttt{SemanticRelation}, categorises the pairing as Added Emoji, Same Category (Congruent), or Distinct Category (Divergent). The random effects structure included random intercepts for speakers and utterances, as well as by-speaker random slopes for \texttt{SemanticRelation} to capture individual variability in prosodic strategies.

\subsection{Results}

\begin{figure*}[t]
\centering
  \includegraphics[width=16cm]{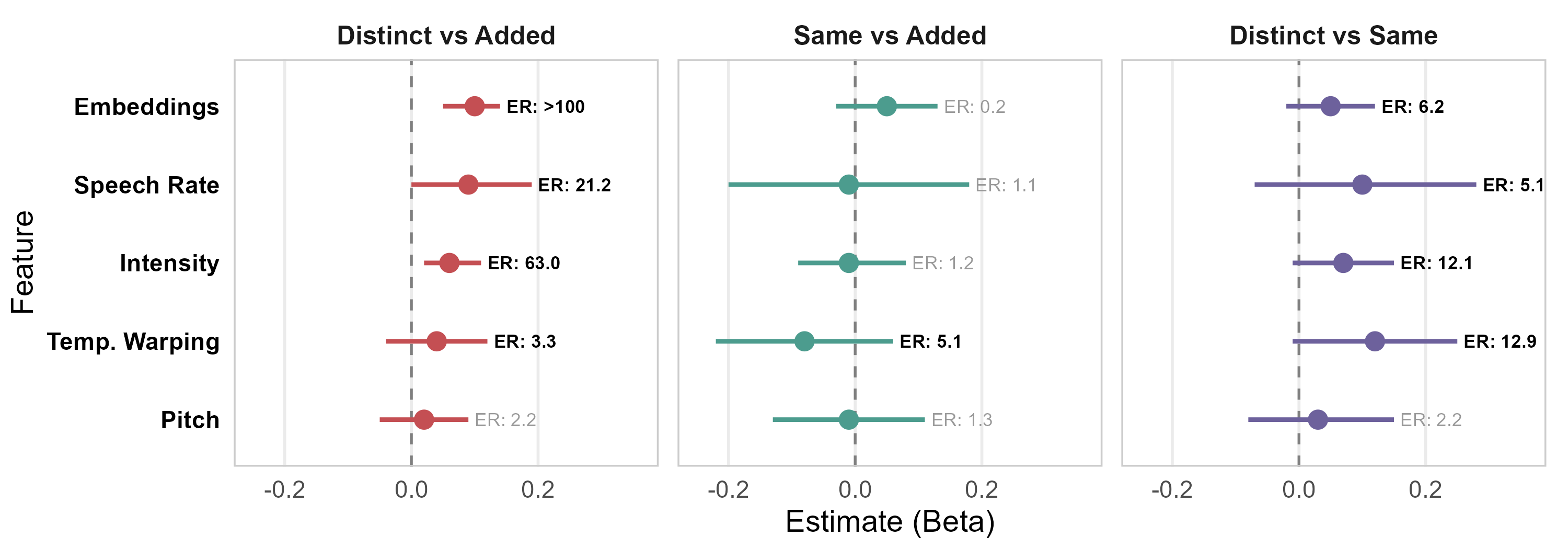}
  \caption{Forest plot of posterior estimates for prosodic shifts across semantic conditions (relative to the `Added' condition). Points represent the mean beta estimate, and error bars indicate the 95\% Credible Interval (CI). Intervals excluding zero indicate a credible non-zero effect. Text labels display the Evidence Ratio (ER), with bold text indicating moderate evidence ($ER > 3$).}
  \label{fig:rq4_results}
\end{figure*}

Figure \ref{fig:rq4_results} summarises the posterior estimates and evidence ratios for each acoustic feature. Given the high inter-speaker variability inherent in prosodic realisation, we provide both 95\% Credible Intervals (CI) and Evidence Ratios (ER) to distinguish between certainty of effect magnitude and probability of effect direction. Following Jeffreys' scale \cite{jeffreys1998theory}, we interpret an ER $> 3$ as “moderate” and $> 10$ as “strong” evidence for a directional shift, allowing for the identification of consistent trends in cases where high variance causes the 95\% CI to overlap zero.

We observed a clear hierarchy where prosodic divergence was largest for Distinct Category pairings, while Same Category pairings clustered near the Added Emoji condition. This “conflict effect” was supported by very strong directional evidence in the Prosodic Embeddings ($\beta=0.10, \text{ER}>100$), driven by a robust increase in Intensity ($\beta=0.06, \text{ER}=63.0$) and a high directional probability for increased Speech Rate ($\text{ER}=21.2$) relative to the Added Emoji condition.
Crucially, speakers also differentiated conflict from congruence: Distinct pairs exhibited greater Temporal Warping ($\text{ER}=12.9$) and Intensity ($\text{ER}=12.1$) than Same pairs, indicating moderate to strong directional evidence for prosodic differentiation. Conversely, synonymous emojis functioned as prosodic stabilisers, showing negligible deviation from the underlying lexical prosody ($\text{ER} \approx 1.2$) and even a potential reduction in temporal warping ($\beta=-0.08, \text{ER}=5.1$). 
Notably, while raw Pitch Distance showed the expected directionality ($Distinct > Added$), the evidence was inconclusive due to high variance ($\text{ER} \approx 2.2$). This stands in contrast to Intensity, which provided a robust univariate signal, and the Prosodic Embeddings, which showed very strong differentiation. This discrepancy likely stems from the acoustic fragility of raw pitch tracking: features like creaky voice or irregular phonation, which are common in expressive speech, introduce noise that obscures the DTW signal.

\section{Discussion}

\subsection{Prosodic Differentiation Across Emojis}
\begin{table}[t]
\centering
\resizebox{\columnwidth}{!}{

\begin{tabular}{lccc}
\toprule
\textbf{Feature} & \textbf{\tear - \pleading} & \textbf{\vomiting - \pleading} & \textbf{\tear - \vomiting} \\\midrule
Embeddings    & 6.03 & 6.14 & \bfseries 6.18 \\
Intensity     & \textbf{0.58} &\textbf{ 0.58} & 0.53 \\

Speech Rate   & 1.28 & \bfseries 1.68 & 0.58 \\
Temp. Warp       & 4.38 & \bfseries 4.48 & 4.45 \\
Pitch         & 1.49 & \bfseries 2.67 & 2.60 \\

\bottomrule
\end{tabular}}
\caption{Prosodic distances between three recordings of the sentence \textit{“The text ‘I miss you’ got me dying”}, spoken by a single participant with different emoji cues. Bolded values indicate the highest distance within each row.}
\label{tab:prosodic-distances}
\end{table}


Figure \ref{fig:example} illustrates prosodic modulation (RQ1) for the sentence \textit{“The text ‘I miss you’ got me dying”} across three emoji contexts.

The \tear and \pleading versions exhibit comparable high, sustained pitch contours, though \pleading is more compressed in duration. In contrast, the \vomiting version diverges with a rising pitch on \textit{“you”} and a sarcastic profile reinforced by a loudness spike on\textit{ “got”}. These observations align with the log-transformed distances in Table \ref{tab:prosodic-distances}. The semantically similar \tear–\pleading pair shows minimal prosodic distance, while the \vomiting–\pleading pair exhibits the highest divergence in temporal alignment and pitch contours. These results align with the broader hierarchy identified in Model 4 regarding the overall magnitude of divergence. While the general model indicates that distinct categories drive larger shifts globally, this specific case illustrates the divergence primarily through pitch and timing.

\subsection{Semantic Overlap and Expressive Limitation}
While many speakers modulate prosody in response to emojis, others produce similar patterns across different emoji conditions. These overlaps raise questions about whether they stem from meaningful semantic similarities or individual expressive limitations, a distinction that is particularly challenging when lexical content implies a dominant tone.

Semantic redundancy often explains this lack of contrast. For instance, in utterances like \textit{“Happy birthday, have a great day”} or \textit{“He started salsa dancing,” }the inherent positivity of the text aligns closely with emojis such as \grinningsmile, \hearteyes, or \dancer. In these cases, the lack of prosodic differentiation likely reflects shared affective intent or semantic alignment rather than a failure to encode the cue. Conversely, other examples suggest expressive constraints. The sentence \textit{“Welp, this is gonna hurt to watch”} was read in an emotionally flat manner for both \grimacing and \sad despite their divergent tones of nervous tension and sadness. Such instances may reflect a difficulty or unwillingness to encode nuanced emotional distinctions in prosody, or a mismatch between the emoji and the speaker’s personal interpretation.

Together, these examples highlight a central challenge for RQ1: similar prosody across conditions may reflect semantic overlap, speaker limitations, or the deliberate downplaying of emoji cues. Disentangling these factors is methodologically difficult and would require detailed manual inspection alongside speaker-level analysis.

\subsection{Prosodic Divergence and Interpretation Failure}

    
Finally, we consider a case where prosodic strategies fail to support accurate listener interpretation (RQ2/RQ3) involving the sentence \textit{“I’m not surprised \scream”}, where \scream is the intended emoji and \vomiting the random alternative in Experiment 4. 

The first speaker produced a lowering pitch contour with a subtle lengthening on “surprised”, giving the utterance a flat, resigned tone signalling emotional detachment rather than shock. Post-hoc interpretation revealed the speaker viewed the emoji as representing “shame,” indicating a misalignment between their internal representation of \scream and its conventional use. The second speaker used a faster tempo and sharper accents that were perceived as sarcastic rather than alarmed. Listeners in both cases favoured the \vomiting emoji, likely due to the mismatch between these prosodic cues and the semantics of the target emoji.

This example highlights several key challenges. Prosodic strategies vary widely between speakers and may not always align with shared semantic expectations. Furthermore, listener interpretation depends not only on prosodic cues but also on culturally or contextually anchored understandings of emoji meaning. Even when prosody is used expressively, its communicative intent can remain opaque without a clear shared mapping between prosodic form and emoji semantics. These failures reinforce our Model 3 findings, demonstrating that successful communication is contingent on speakers and listeners aligning their paralinguistic intuitions.

\section{Conclusion}

This work provides the first systematic evidence linking emoji semantics to prosodic expression and interpretation in speech. We establish that emojis act as mediators of prosodic intent, allowing speakers to encode pragmatic nuance into otherwise static text. Our results reveal a hierarchy in this process: the magnitude of prosodic divergence is systematically governed by the semantic distance between emojis, with intensity and timing serving as primary acoustic carriers of emoji-driven intent. Finally, we demonstrate that these modulations enable listeners to recover intended meanings with accuracy significantly above chance.
 
Beyond these findings, this study highlights emojis as a structured paralinguistic layer bridging the gap between digital text and spoken production. Future work can leverage these findings to ensure that multimodal large language models (LLMs) treat emojis as dynamic suprasegmental operators rather than simple text tokens. Furthermore, this foundation paves the way for TTS systems and screen readers capable of conveying authentic pragmatic intent, ultimately positioning emojis as tools for modelling the nuances of human communication.

\section*{Limitations}
While our empirical analysis establishes a robust baseline for emoji-driven prosody, this study represents an exploratory first step rather than an exhaustive account. First, we prioritised lexical control via elicited reading over spontaneous conversation. Although participants received no instructions to focus on emojis, the resulting prosody may be more deliberate than in casual speech. Second, untrained speakers exhibited significant variation in expressivity. While our Bayesian models accounted for individual baselines, we observed that the clarity of prosodic cues is graded and varies by speaker, rather than being uniform across the population. Relatedly, our dataset is restricted to British English speakers, and we did not control for specific regional dialects during recruitment. Although our speaker-level random intercepts help mitigate baseline production variability, future work should consider finer-grained sociolinguistic factors. Third, the inter-speaker analysis in Experiment 4 relies on a small listener sample size. While sufficient for initial model convergence, this represents a preliminary exploration of prosodic convergence that requires scaling up in future work. Fourth, we acknowledge that automated prosodic distance metrics such as DTW on pitch and intensity are imperfect proxies for human pragmatic perception and are inherently vulnerable to acoustic fragilities. Finally, our curated stimuli may introduce selection bias; we did not control for cross-platform emoji rendering, and our findings may not generalise to languages with distinct prosodic or pragmatic norms.

\section*{Ethical Considerations}

This research was granted ethics approval by the Informatics Ethics Committee, University of Edinburgh (Application Number: 243719), with all participants providing explicit informed consent for the use and release of their recordings and annotations for research purposes. Recruitment was conducted via Prolific and restricted to UK residents. Participants were paid £1.50 per task (pro-rated for duration), resulting in an effective compensation rate of at least £9/hour, adhering to Prolific’s “Ethical Rewards” guidelines. To maintain privacy, data was anonymised at the source using Prolific IDs, ensuring no direct identifiers were collected. To verify data integrity and safety, we utilised an Automatic Speech Recognition (ASR) model (Whisper) to ensure recordings matched the target stimuli and contained no personally identifying information. This automated process was supplemented by manual audits of a subset of recordings and annotations to verify the absence of harmful content.

\section*{Acknowledgments}
This work was supported in part by the Centre for Doctoral Training in Natural Language Processing, funded by UKRI (grant EP/S022481/1). We extend our sincere thanks to Catherine Lai and Artemis Deligianni (University of Edinburgh) for their valuable advice and insights regarding the experimental design and statistical modelling.

\bibliography{custom, anthology}

\appendix
\section{Model Specifications}\label{app:models}

All models were implemented in R \citep[][version 4.5.2]{R} using the \texttt{brms} package \cite{brms}, which interfaces with the Stan probabilistic programming platform. To ensure rigorous confirmatory analysis, we did not perform post-hoc model comparisons (e.g., likelihood ratio tests) to select predictors. Instead, we determined our model structures and likelihood functions \textit{a priori} based strictly on the experimental design. Binary outcomes for interpretation and perception trials (RQ1, RQ2, and RQ3) were modelled using a Bernoulli distribution with a logit link function. Continuous prosodic distance measures (RQ4) were modelled using Gaussian distributions; to satisfy normality assumptions for these continuous models, the distance metrics were log-transformed and Z-standardised prior to fitting (see Section 5.2).

We applied weakly informative priors to the fixed effects ($Normal(0, 0.5)$) to provide regularisation and ensure stable posterior convergence. This prior choice reflects a conservative assumption that large effect sizes are relatively unlikely, helping prevent the models from over-fitting to noise in the high-variance speech data. To minimise Type I errors, we followed the ``Keep It Maximal'' principle \cite{barr2013random} for specifying random effects. For our continuous acoustic analysis (Model 4), we successfully fitted a maximal structure that included random slopes for semantic conditions by speaker (e.g., \texttt{1 + SemanticRelation | Speaker}). This accounts for individual variability in how speakers adapt their prosodic strategies to different emoji cues, ensuring our population-level estimates are robust to speaker-level idiosyncratic behaviour. However, for the binary perception tasks (Models 1--3), maximal structures failed to converge, a common issue with sparse binomial data. We therefore simplified these models to random-intercept-only structures to ensure stable and reliable parameter estimation. Finally, for all Bayesian models, we rigorously assessed chain convergence. We verified that the Gelman-Rubin convergence diagnostic ($\hat{R}$) was strictly less than 1.01 for all parameters, visually inspected trace plots for healthy mixing, and ensured sufficient Effective Sample Size (ESS) for stable posterior inference.

\section{Data statistics}\label{app:data}

\subsection{Listener Accuracy - Intra Speaker}\label{app:data3}
\begin{figure}[t]
\centering
  \includegraphics[width=\linewidth]{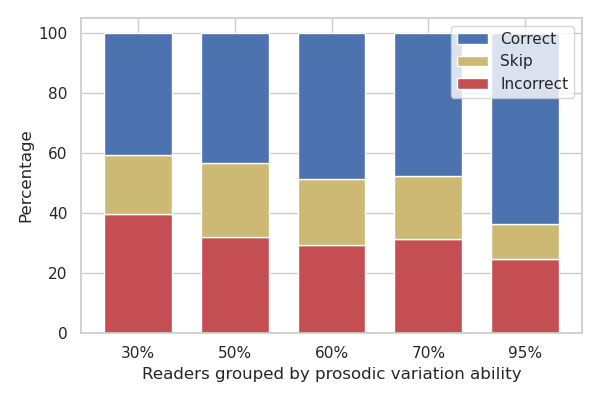}
  \caption{Listener (Experiment 3) accuracy in emoji assignment for prosodically distinct pairs, grouped by speaker variation. Bars show correct, incorrect, and skipped responses from least to most prosodically varied speakers. }
  \label{fig:ex3_results}
\end{figure}

Figure \ref{fig:ex3_results} illustrates how a speaker’s expressiveness directly impacts communicative success. A Spearman’s rank correlation revealed a significant positive relationship between speaker variation and correct emoji identification ($\rho = 0.90, p = .037$) and a significant negative correlation for incorrect assignments ($\rho = -0.90, p = .037$). For the most expressive speakers (95\% bin), accuracy reached 62.50\%, significantly exceeding both the incorrect rate (25.52\%) and a random-choice baseline. Conversely, in the lowest variation group (30\%), correct (41.09\%) and incorrect (39.11\%) rates were nearly equal, indicating that listeners were reduced to random guessing. While skip rates dropped from 20\% to 11.98\% in the highest bin, this trend was not statistically significant ($\rho = -0.10, p = .87$).

\subsection{Listener Accuracy - Inter Speaker}\label{app:data4}

\begin{table}[t]
\centering
\resizebox{\columnwidth}{!}{
\begin{tabular}{l|c|ccc}
\toprule 
\textbf{Response} & \textbf{Overall} & \textbf{Correct} & \textbf{Partial} & \textbf{Incorrect} \\
\midrule
\textbf{Same}      & 34.74\% & 66.67\% & 6.06\%  & 27.27\% \\
\textbf{Different} & 65.26\% & 14.75\% & 74.59\% & 10.66\% \\
\bottomrule
\end{tabular}}
\caption{Results from the inter-speaker experiment showing how listeners judge prosodic similarity between two recordings and how these judgments affect their accuracy in assigning each recording to the correct emoji.}
\label{tab:inter-speaker}
\end{table}

Results in Table \ref{tab:inter-speaker} show that 34.74\% of inter-speaker pairs were judged as sounding similar, while 65.26\% were judged as different. This suggests that speakers often employ flexible prosodic strategies for the same emoji. In this experiment, listeners were presented with two recordings and two emoji options, allowing for four possible assignment combinations. We define Correct as the proper identification of the target emoji for both recordings, while Partial correctness occurs when a listener assigns the target emoji to only one of the two recordings. Incorrect reflects a failure to identify the target emoji in either instance. When listeners perceived prosodic similarity, they achieved full correctness 66.67\% of the time, indicating strong interpretive agreement when prosody converges. For pairs judged as different, the high partial correctness rate (74.59\%) reflects increased uncertainty; listeners often successfully “recovered” the emoji for one speaker while failing for the other. However, even when prosody differed, the full correctness rate (14.75\%) still exceeded the incorrect rate (10.66\%), suggesting that divergent realisations can still convey intended meaning, albeit with less reliability.

\section{Analysis of Perceptual Overlap - RQ2}
\begin{table}[t]
\centering
\resizebox{\columnwidth}{!}{
\begin{tabular}{lrc}
\toprule
\textbf{Predictor} & \textbf{Est. ($\beta$)} & \textbf{95\% CI} \\
\midrule
(Intercept)                       & -1.87 & $[-2.28, -1.49]$ \\
Speaker Expressivity              & -0.18 & $[-0.37, 0.00]$ \\
Contrast Type (Emoji--NoEmoji)            & 0.120  & $[-0.20, 0.59]$ \\
Speaker Emoji Use                 & -0.05 & $[-0.23, 0.14]$ \\
Listener Emoji Use                & 0.15  & $[-0.18, 0.48]$ \\
Speaker $\times$ Listener Use     & -0.06 & $[-0.26, 0.13]$ \\
\bottomrule
\end{tabular}}
\caption{Bayesian multilevel logistic regression results predicting the probability of a “semantic tie” (assigning the same emoji to both recordings)}
\label{tab:semantic_ties}
\end{table}
To further investigate the mechanics of communicative success, we analysed the likelihood of “semantic ties.” These represent instances where a listener assigned the same emoji to both recordings within a single trial, modelled as a “Skip” in Figure \ref{fig:ex3_results}. A high rate of semantic ties indicates perceptual overlap, where the speaker's prosodic modulation is insufficient to signal a change in symbolic intent. We modelled the probability of such ties using a Bayesian multilevel logistic regression:\begin{center} \textbf{Model 5}: \texttt{SemanticTie $\sim$ SpeakerExpressivity + ContrastType + (SpeakerEmojiUse $\times$ ListenerEmojiUse) + (1 | Utterance) + (1 | Listener)}\end{center}The results in Table \ref{tab:semantic_ties} show a strongly negative intercept ($\beta = -1.87, CI [-2.28, -1.49]$), indicating that the baseline probability of a semantic tie is very low ($P \approx 0.13$). This suggests that listeners are naturally predisposed to seek out prosodic distinctions rather than perceiving the recordings as identical. The primary driver of further reducing this perceptual ambiguity was Speaker Expressivity ($\beta = -0.18, CI [-0.37, 0.003]$). While the credible interval narrowly overlaps zero, the negative trend suggests that higher expressive bandwidth effectively helps listeners to recognise distinct meanings. By producing more distinctive prosodic signals, highly expressive speakers successfully disambiguate the potential semantic tie, making it easier for listeners to identify that two different meanings were intended. This finding reinforces the conclusion that the interpretability of emoji-enriched speech is directly contingent on the speaker's ability to vocalise pragmatic contrasts.

\section{Listener Accuracy and Emoji Ambiguity}\label{app:ambiguity}

\begin{table}[t]
\centering
\resizebox{\columnwidth}{!}{

\begin{tabular}{ccccc}
\toprule
 & \textbf{Ambiguity} & \textbf{Correct} & \textbf{Partial} & \textbf{Incorrect} \\
\midrule
\question & 0.158 & 25.0  & 75.0  & 0.0  \\
\wink & 0.171 & 83.3  & 16.7  & 0.0  \\
\fist & 0.213 & 0.0  & 100.0 & 0.0 \\
\tear & 0.244 & 66.7  & 33.3  & 0.0  \\
\scream & 0.375 & 0.0  & 20.0 & 80.0 \\
\confused & 0.463 & 100.0 & 0.0 & 0.0 \\
\smirk & 0.516 & 66.7 & 0.0  & 33.3 \\
\nomouth & 0.541 & 22.2 & 44.4 & 33.3 \\
\thumbsup & 0.659 & 0.0  & 50.0 & 50.0 \\

\bottomrule
\end{tabular}}
\caption{Examples illustrating emoji ambiguity and Experiment 4 listener interpretation accuracy (in percentages), sorted from low to high ambiguity.}
\label{tab:emoji-ambiguity-examples}
\end{table}

Table \ref{tab:emoji-ambiguity-examples} presents a subset of emojis from Experiment 4 alongside listener interpretation accuracy and ambiguity scores derived from the semantic variation metrics reported by \citet{czkestochowska2022context}. These scores reflect cross-participant variability in emoji interpretation when presented without context, where higher values indicate greater semantic ambiguity. Although the sample size precludes statistical generalisation, the examples illustrate a qualitative trend: emojis with lower ambiguity scores tend to yield fewer incorrect interpretations, whereas highly ambiguous emojis result in greater listener disagreement.

Specific emojis revealed distinct patterns within the data. Emojis such as \raisehand and \confused demonstrated clear prosodic convergence alongside high accuracy. In contrast, emojis like \thinking and \hug were reliably recognised despite divergent prosody, suggesting that inherent semantic context may at times override prosodic mismatch. Conversely, for \joy, \scream, and \nomouth, prosodic similarity did not consistently ensure accurate interpretation.

Qualitative comparison with existing ambiguity scores suggests that while semantic ambiguity does not dictate the degree of prosodic convergence, emojis with lower inherent ambiguity tend to result in fewer incorrect interpretations. These tentative patterns suggest that higher semantic ambiguity may reduce the recoverability of emoji intent from prosody alone, as listeners lack a stable semantic “anchor” to which they can map vocal variations.

\section{Emoji Meaning Analysis}\label{app:embedding}




To analyse the semantic consistency of emoji usage, we compared three meaning representations for each emoji: the \textit{intent} (from Experiment~1), the \textit{interpretation} (from Experiment~2), and the official Unicode \textit{description}.

For each emoji, we aggregated the most frequent annotations and merged them into a representative string per meaning type. We then generated sentence embeddings using the \texttt{sentence-transformers/all-mpnet-base-v2} model \cite{reimers-gurevych-2019-sentence, song2020mpnet}. This yielded one embedding per emoji for each semantic category.

We applied UMAP \cite{mcinnes2018umap-software} to reduce the embeddings to two dimensions for visualisation. Emojis were plotted directly in 2D space using their glyphs, after normalising representations (e.g., removing skin tone and gender modifiers) to avoid redundancy and improve clarity (Figure \ref{fig:emoji_embeddings_ex2}).

\begin{figure}[t]
\centering
  \includegraphics[width=\linewidth]{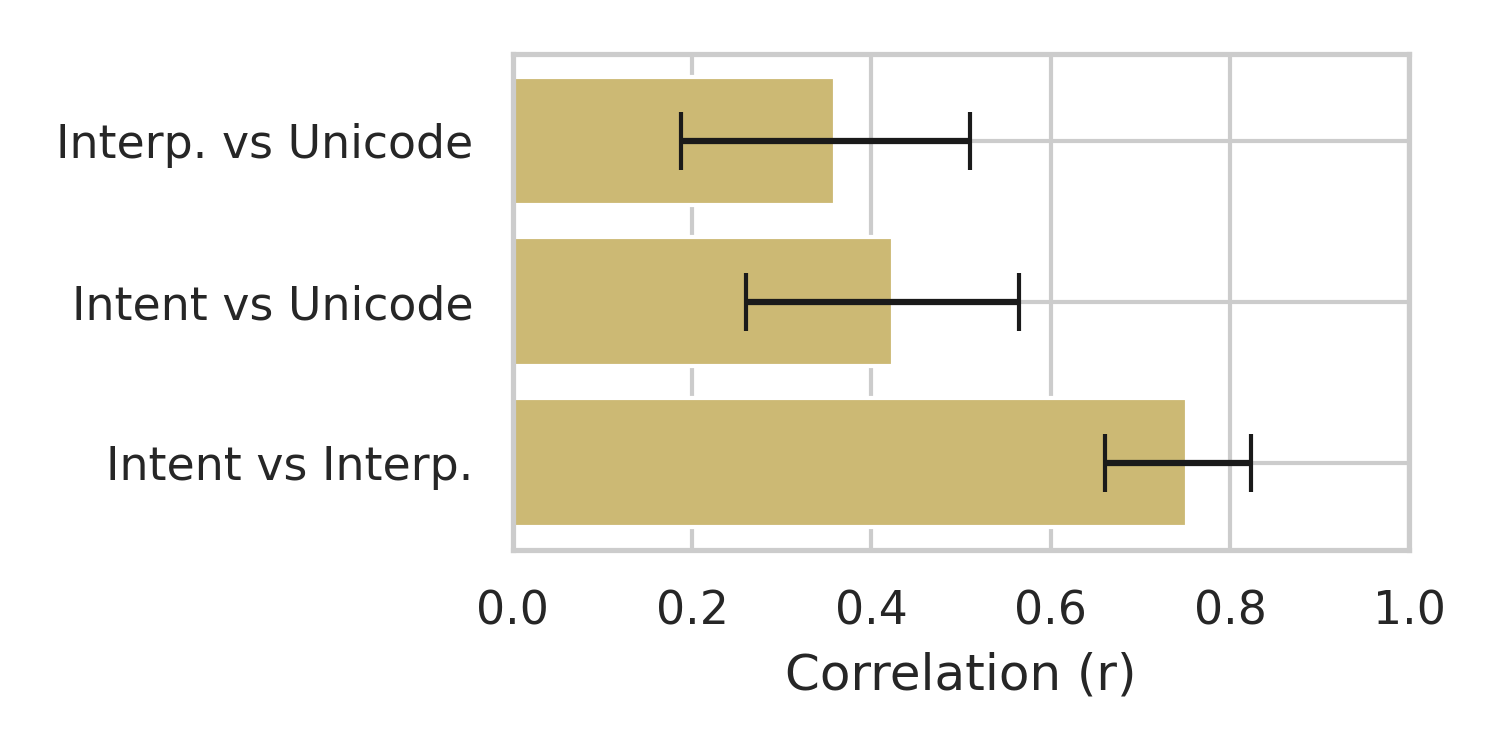}
  \caption{Mean Pearson correlations between emoji meaning spaces (Intent vs Interpretation vs Unicode Descriptions). Error bars indicate 95\% confidence intervals computed using Fisher z-transformation.}
  \label{fig:correlation}
\end{figure}

We evaluated the alignment between the three emoji meaning spaces by computing pairwise Pearson correlation coefficients between corresponding emoji embeddings. To assess the reliability of these correlations across the dataset, we applied the Fisher z-transformation, which stabilises the variance of correlations and enables us to compute statistically sound confidence intervals for the mean. Figure~\ref{fig:correlation} shows the mean correlation and associated 95\% confidence intervals across all emojis with at least five annotations. We observe the strongest alignment between intent and interpretation embeddings (\textbf{mean $r$ = 0.752}, CI: [0.660, 0.823]), reflecting high agreement between what users intend and how others interpret emoji use. In contrast, the unicode description space shows significantly weaker alignment with both intent (\textbf{$r$ = 0.424}, CI: [0.260, 0.564]) and interpretation (\textbf{$r$ = 0.359}, CI: [0.188, 0.510]), reinforcing prior findings that Unicode definitions often fail to capture how emojis are actually used in communicative context.


\section{Emoji Clustering} \label{app:clustering}

To explore emergent semantic groupings among emojis, we performed unsupervised clustering on the embeddings derived from Experiments~1 and 2, as well as from the emoji Unicode descriptions. We first applied Principal Component Analysis (PCA) to determine the intrinsic dimensionality of the embedding space. Based on the explained variance, we selected 49 dimensions to preserve approximately 90\% of the variance. 

Clustering was performed using HDBSCAN \cite[Hierarchical Density-Based Spatial Clustering of Applications with Noise]{campello2013density}, a density-based algorithm that can identify an optimal number of clusters based on the density distribution of the data. HDBSCAN does not require the number of clusters to be predefined and is robust to noise and outliers. Emojis that did not belong to any cluster were labelled as noise.  We set \texttt{min\_cluster\_size} to 3 to allow for finer-grained clusters, and used Euclidean distance as the similarity metric.

Figure \ref{fig:emoji_embeddings_ex2} shows the two-dimensional projections of the resulting embeddings and clusters. The figure shows labels for each cluster derived from the most frequent emoji interpretations within the group. These derived clusters serve as our semantic categories in subsequent analyses.

To assess the alignment between emoji clusters derived from different semantic spaces (Intent, Interpretation, and Description), we computed the \textit{Adjusted Rand Index} \cite[ARI]{hubert1985comparing} and \textit{Normalized Mutual Information} \cite[NMI]{JMLR:v11:vinh10a} between each pair of clusterings. These metrics quantify the similarity of cluster assignments while accounting for chance (ARI) and shared information (NMI).

\begin{table}[t]
\centering
\begin{tabular}{lcc}
 \toprule
\textbf{Cluster Comparison} & \textbf{ARI} & \textbf{NMI} \\ \midrule
Int. vs Inter.        & 0.249 & 0.613 \\  
Int. vs Desc.         & 0.077 & 0.431 \\  
Inter. vs Desc. & 0.062 & 0.443 \\  
\bottomrule
\end{tabular}
\caption{Adjusted Rand Index (ARI) and Normalized Mutual Information (NMI) between clusterings based on different semantic spaces.}
\end{table}

The ARI scores suggest moderate agreement between \textit{Intent} and \textit{Interpretation} clusters ($\text{ARI}=0.249$), whereas clusterings involving \textit{Description} show weak alignment (ARI $< 0.1$). NMI results are consistent with this pattern: the Intent and Interpretation spaces share more information ($\text{NMI}=0.613$) than either does with the Description space ($\text{NMI} \approx 0.43$). These findings indicate that user-generated intentions and interpretations exhibit a shared semantic structure that is not well captured by official Unicode descriptions.

\section{Instructions and Trial Samples}\label{app:instructions}
\subsection{Experiment 1}
Participants were given the following instructions:“
\textit{
\begin{enumerate}
    \item Select ONE emoji and place it in any suitable position within the provided sentence. Only use emojis and spaces, avoiding inserting any other elements.
    \item Describe how the chosen emoji modifies the sentence. Keep it short! One word should be enough.
    \item Repeat until you have filled out at \textbf{least two forms} with a suitable emoji and its corresponding descriptions.
\end{enumerate}
Avoid choosing multiple emojis that are too similar.}

\textit{For instance, if you select a happy emoji, consider avoiding another happy emoji to add diversity to your selections.”}

Figure \ref{fig:exp1_example} shows an example of a trial page for Experiment 1.
\subsection{Experiment 2}
Participants were asked to complete the Recording Tasks first, and then provide their interpretation of the encountered emojis. Here, the instructions given to the participants.

\subsubsection{Recording Task}

\textit{“Click on the mic icon to start recording audio, then click again to stop.
}

\textit{You will be able to playback your recording. You can confirm by pressing 'Continue' or try again by pressing 'Record again' (it will start recording immediately).
}

\textit{Try to record in a quiet environment.”}

Figure \ref{fig:exp2_example} shows the recording interface.

\subsubsection{Annotation Task}

\textit{“In this second (and last) part, you will be asked to describe how you interpreted the emoji used in the previous section.
}

\textit{Keep it short!  \textbf{One word} should be enough.
}

\textit{Leave blank if there wasn't any emoji.”
}
\subsection{Experiment 3}

\begin{figure*}[t]
  \centering

  \begin{subfigure}{\linewidth}
    \centering
    \includegraphics[width=0.9\linewidth]{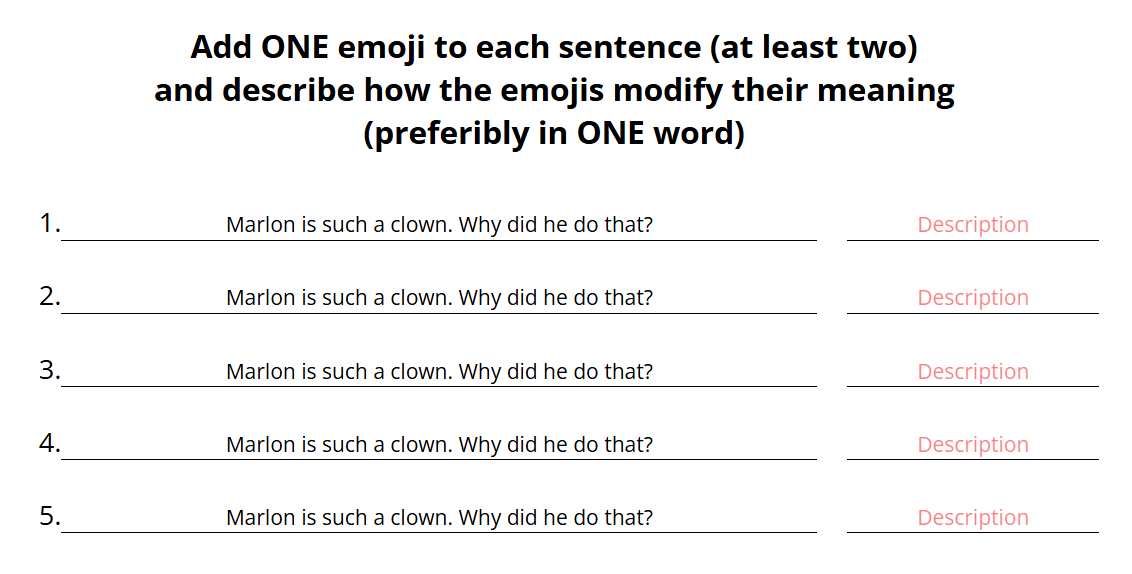}
    \caption{Experiment 1 - Emoji Annotation Task}
    \label{fig:exp1_example}
  \end{subfigure}  
  
  \begin{subfigure}{\linewidth}
    \centering
    \includegraphics[width=0.6\linewidth]{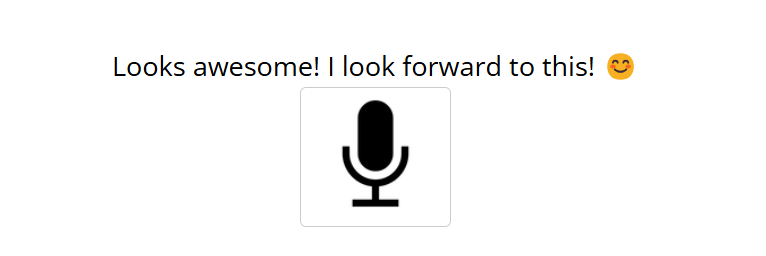}
    \caption{Experiment 2 - Recording Task}
    \label{fig:exp2_example}
  \end{subfigure}

  \begin{subfigure}{\linewidth}
    \centering
    \includegraphics[width=0.9\linewidth]{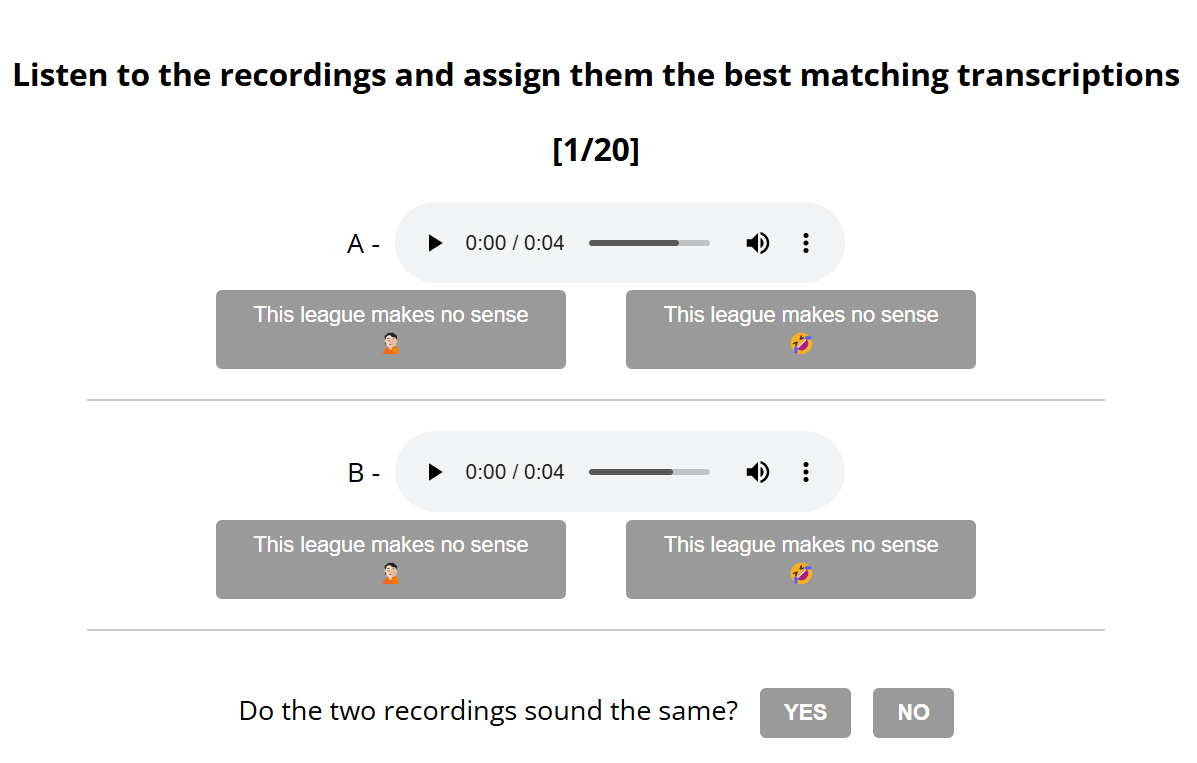}
    \caption{Experiment 3 and 4 - Listening Task}
    \label{fig:exp3_example}
  \end{subfigure}  
  
  \caption{Example of trials' main page for the online experiments.}  
  \label{fig:exp_example}
\end{figure*}

Figure \ref{fig:exp3_example} shows an example of the interface for Experiment 3. The same interface was presented for experiments with Inter and Intra-Speaker Data.

\subsubsection{Inter-Speaker Data}

\textit{“In this study, you will listen to pairs of recorded speech. Each pair will contain the same words, but they may differ in how they are spoken. For example, in rhythm, intonation, intensity, use of pauses and so on. Your task is to determine whether there are intentional variations between the two recordings. Please disregard any differences that do not seem deliberate. For example, small lexical variations or stutter.
}

\textit{You will be provided with transcriptions of the recordings, which may contain different emojis or none at all. Your task is to match each recording to the transcription that best corresponds to it. 
}

\textit{Try to assign each recording to a different transcription first!”
}

\subsubsection{Intra-Speaker Data}

\textit{“In this study, you will listen to pairs of recorded speech from \textbf{two different speakers}. Each pair will contain the \textbf{same words, but they may differ in how they are spoken}, for example, in rhythm, intonation, intensity, use of pauses and so on. Your task is to determine whether there are intentional variations between the two recordings. Please disregard any differences that do not seem deliberate. For example, small lexical variations or stutter.
}

\textit{You will hear \textbf{two recordings of the same sentence, spoken by different speakers. }
}

\textit{Your task is to listen carefully to each recording and choose the emoji that best matches its tone.
}

\textit{Each speaker produced their recording based on an emoji prompt.
}

\textit{You can assign the \textbf{same emoji to both recordings}, or a \textbf{different one to each}, depending on what you perceive.”
}

\end{document}